\documentclass[twoside,11pt]{article}

\usepackage{blindtext}
\usepackage{xspace}
\usepackage{makecell}
\usepackage{amsmath}
\usepackage{multirow}
\usepackage{booktabs}
\usepackage{listings}
\usepackage{xcolor}
\usepackage{subcaption}
\usepackage{enumitem}
\usepackage{rotating}

\definecolor{codegreen}{rgb}{0,0.6,0}
\definecolor{codegray}{rgb}{0.5,0.5,0.5}
\definecolor{codepurple}{rgb}{0.58,0,0.82}
\definecolor{backcolour}{rgb}{0.95,0.95,0.92}

\usepackage{jmlr2e}

\usepackage{todonotes}
\usepackage{xcolor}
\newif{\ifhidecomments}
\hidecommentsfalse
\ifhidecomments
\newcommand{\wjdd}[1]{\todo[linecolor=cyan,backgroundcolor=cyan!25,bordercolor=cyan,size=\scriptsize]{}
\newcommand{\wjd}[1]{{\color{cyan}{}
\else
\newcommand{\wjdd}[1]{\todo[linecolor=cyan,backgroundcolor=cyan!25,bordercolor=cyan,size=\scriptsize]{(Jindong) #1}}
\newcommand{\wjd}[1]{{\color{cyan}{[(Jindong) #1]}}}
\fi

\newcommand{\dataset}{{\cal D}}
\newcommand{\fracpartial}[2]{\frac{\partial #1}{\partial  #2}}
\newcommand{\method}{PromptBench\xspace}
\newcommand{\prompt}[1]{{\ttfamily #1}\xspace}

\newcommand{\rone}[1]{{\color{black}{#1}}}
\newcommand{\rtwo}[1]{{\color{black}{#1}}}
\newcommand{\rthree}[1]{{\color{black}{#1}}}
\newcommand{\ronertwo}[1]{{\color{black}{#1}}}
\newcommand{\ronerthree}[1]{{\color{black}{#1}}}
\newcommand{\rtworthree}[1]{{\color{black}{#1}}}
\newcommand{\ronertworthree}[1]{{\color{black}{#1}}}

\newcommand{\glue}{GLUE\xspace}
\newcommand{\sst}{SST-2\xspace}
\newcommand{\cola}{CoLA\xspace}
\newcommand{\mnli}{MNLI\xspace}
\newcommand{\qnli}{QNLI\xspace}
\newcommand{\rte}{RTE\xspace}
\newcommand{\wnli}{WNLI\xspace}
\newcommand{\qqp}{QQP\xspace}
\newcommand{\mrpc}{MRPC\xspace}
\newcommand{\mmlu}{MMLU\xspace}
\newcommand{\squad}{SQUAD V2\xspace}
\newcommand{\un}{UN Multi\xspace}
\newcommand{\iwslt}{IWSLT 2017\xspace}

\newcommand{\dolly}{Dolly\xspace}
\newcommand{\celebras}{Celebras\xspace}
\newcommand{\llama}{LLaMa2\xspace}
\newcommand{\vicuna}{Vicuna\xspace}
\newcommand{\nexo}{Nexo\xspace}
\newcommand{\chat}{ChatGPT\xspace}
\newcommand{\llms}{LLMs\xspace}

\newcommand{\totaldataset}{$22$\xspace}
\newcommand{\totaltask}{$12$\xspace}
\newcommand{\totalattack}{$7$\xspace}
\newcommand{\totalmodel}{$9$\xspace}

\newcommand{\chh}[1]{\todo[linecolor=blue,backgroundcolor=blue!25,bordercolor=blue,size=\scriptsize]{(CH): #1}}
\newcommand{\ch}[1]{{\color{blue}{[(CH): #1]}}}

\usepackage{lastpage}
\jmlrheading{25}{2024}{1-\pageref{LastPage}}{1/24; Revised
6/24}{8/24}{24-0023}{Kaijie Zhu, Qinlin Zhao, Hao Chen, Jindong Wang, Xing Xie}

\ShortHeadings{PromptBench: A Unified Library for Evaluation of Large Language Models}{Zhu, Zhao, Chen, Wang, Xie}
\firstpageno{1}

\begin{document}

\title{\method: A Unified Library for Evaluation of Large Language Models}

\author{Kaijie Zhu$^{1,2}$\thanks{The first two authors contributed equally. Work done at MSRA.}, Qinlin Zhao$^{1,3\ast}$, Hao Chen$^{4}$, Jindong Wang$^{1}$\thanks{Corresponding author: Jindong Wang (jindong.wang@microsoft.com).}, Xing Xie$^{1}$ \\ 
\normalfont{$^1$Microsoft Research Asia \quad $^2$Institute of Automation, Chinese Academy of Sciences\\ $^3$University of Science and Technology of China \quad $^4$Carnegie Mellon University}}

\editor{Zeyi Wen}

\maketitle

\begin{abstract}
The evaluation of large language models (\llms) is crucial to assess their performance and mitigate potential security risks.
In this paper, we introduce \method, a unified library to evaluate \llms.
It consists of several key components that can be easily used and extended by researchers: prompt construction, prompt engineering, dataset and model loading, adversarial prompt attack, dynamic evaluation protocols, and analysis tools.
\method is designed as an open, general, and flexible codebase for research purpose. It aims to facilitate original study in creating new benchmarks, deploying downstream applications, and designing new evaluation protocols.
The code is available at: \url{https://github.com/microsoft/promptbench} and will be continuously supported.
\end{abstract}

\begin{keywords}
  Evaluation, large language models, framework
\end{keywords}

\section{Introduction}
\label{sec:intro}

Large language models (\llms) are revolutionizing aspects of human life and society, \rthree{such as medical diagnostics~\citep{mcduff2023accurate, thirunavukarasu2024ophthalmology}, and educational tools~\citep{ho2023reasoning}.}
Evaluation is of paramount importance to understand the true capabilities of \llms, mitigate potential risks, and eventually, benefit society further~\citep{eisenstein2023test,chang2023survey}.
\rthree{Recent efforts have evaluated \llms from diverse aspects \citep{mmlu, liang2022holistic,zheng2023judging,li2023alpacaeval,huang2023c,leaderboard}.}
Among the findings, one of the most important is that current \llms are sensitive to prompts \citep{wang2023robustness}, vulnerable to adversarial prompt attacks~\citep{zhu2023promptbench}, and exposed to testset data contamination~\citep{willig2023causal,zhou2023don,zhu2023dyval}, which pose severe security and privacy issues~\citep{wang2023decodingtrust,simmons2022moral}.
On top of that, there have been various prompt learning algorithms developed based on different evaluation metrics, such as BDPL~\citep{diao2022black}, GrIPS~\citep{prasad2022grips} and Plum~\citep{pan2023plum}.
Given the increasing popularity of \llms, it is indispensable to develop a unified codebase to enable easy, fast, and flexible evaluation of large foundation models.

There are existing libraries such as LlamaIndex~\citep{llamaindex}, semantic kernel~\citep{semantic_kernel}, and LangChain~\citep{langchain}.
LlamaIndex and LangChain enhance LLM applications by incorporating databases and various data sources. Semantic Kernel aims to merge AI services with programming languages for versatile AI app development.
Eval-harness~\citep{eval-harness} offers a comprehensive framework for evaluating generative language models.
Zeno~\citep{zeno} is an AI evaluation platform supporting interaction and visualization, but it is not easy to customize.
LiteLLM~\citep{litellm} implements a unified API call for different LLM service prodiders.


This paper introduces \textbf{\method}, a unified Python library to evaluate \llms from comprehensive dimensions.
\rthree{
It is designed to fill the gaps current libraries have, offering comprehensive support not only for standard model evaluations but also for advanced scenarios including adversarial prompt attacks and dynamic evaluations. 
Its extensible architecture allows for the incorporation of new evaluation protocols, addressing the limitations found in other tools.
The detailed comparisons are shown in Appendix~\ref{sec-append-related}, highlighting how \method provides a more complete toolkit for the nuanced evaluation of language models, especially in research contexts where adaptability and thoroughness are paramount.}
It consists of a wide range of \llms and evaluation datasets, covering diverse tasks, evaluation protocols, adversarial prompt attacks, and prompt engineering techniques.
As a holistic library, it also supports several analysis tools for interpreting the results.
Our library is designed in a modular fashion, allowing researchers to easily build evaluation pipelines for their own projects.
We open-source \method with comprehensive documents and tutorials\footnote{\url{https://promptbench.readthedocs.io/en/latest/}} to support easy, flexible, and collaborative evaluation.
We believe \method could enhance our understanding of \llms and spur new research within the community.

\section{\method}
\label{sec:method}

\begin{figure}[t!]
\centering
\includegraphics[width=1.0\textwidth]{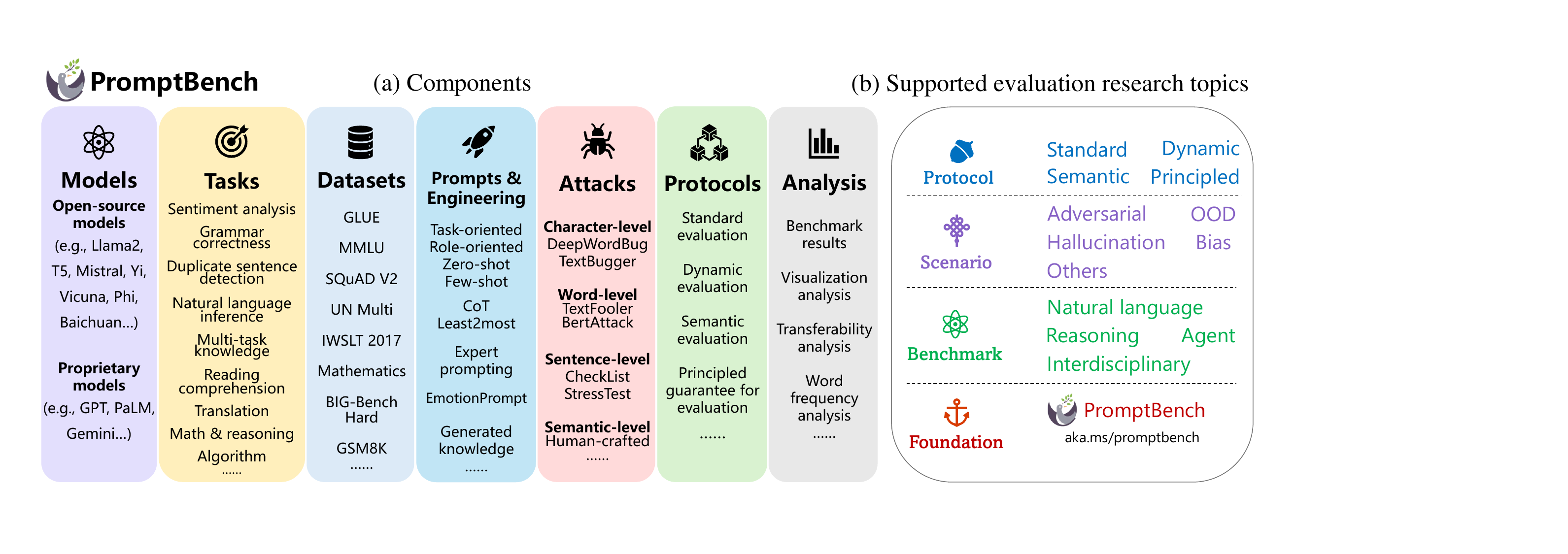}
\vspace{-.2in}
\caption{The components and supported research areas of \method.}
\label{fig-main}
\vspace{-.2in}
\end{figure}

\method can be easily installed either via \prompt{pip install promptbench} or \prompt{git clone}.
In this section, we briefly introduce the components of \method and how to use it to build an evaluation pipeline for \llms. An overview of \method is shown in \figurename~\ref{fig-main}.

\subsection{Components}

\textbf{Models.}
\method supports both open-source and proprietary \llms and \ronertwo{VLMs} and it is open to add more. 
\ronertworthree{Currently, it supports a diverse range of \llms and VLMs, ranging from Llama2 series~\citep{touvron2023llama2}, Mixtral series~\citep{jiang2024mixtral}, LlaVa series~\citep{liu2023llava} to GPT series~\citep{chatgpt, openai2023gpt4}. 
\method provides unified \prompt{LLMModel} and \prompt{VLMModel} interfaces to allow easy construction and inference of a model with specified max generating tokens and generating temperature.} \rone{The interfaces also support customized models, including those that have been fine-tuned for specific applications.}
More details of the supported models are shown in Appendix \ref{sec:appendix-model}. 

\textbf{Datasets and tasks.}
\rthree{PromptBench comprises a wide array of tasks, currently supporting diverse challenges across \totaltask tasks and \totaldataset public datasets, with the capacity for additional expansions. The supported tasks include fundamental NLP tasks such as sentiment analysis, grammar correctness, and duplicate sentence detection, as well as complex challenges involving natural language inference, multi-task knowledge, and reading comprehension. It also covers specialized areas like translation, mathematical problem-solving, and various forms of reasoning—logical, commonsense, symbolic, and algorithmic. For detailed descriptions of each dataset and specific task configurations (see Appendix~\ref{sec:appendix-task}). The unified \prompt{DatasetLoader} interface facilitates easy and customizable loading and processing of these datasets, enhancing the usability and flexibility of PromptBench.
}

\textbf{Prompts and prompt engineering.}
\method offers a suite of $4$ distinct prompt types, and additionally, users have the flexibility to craft custom prompts using the \prompt{Prompt} interface. Task-oriented prompts are structured to clearly delineate the specific task expected of the model, whereas role-oriented prompts position the model in a defined role, such as an expert, advisor, or translator. These prompt categories are adaptable for both zero-shot and few-shot learning contexts, offering diverse application possibilities.
Moreover, \method currently includes $6$ prominent prompt engineering methods, \rthree{details can be found in Appendix~\ref{append:pe}.} Our framework is not only equipped for the easy integration of these existing techniques through the \prompt{prompt\_engineering} module but is also actively evolving to encompass a broader spectrum of prompt engineering methods, enhancing its adaptability in varied evaluation scenarios.

\textbf{Adversarial prompt attacks.}
To facilitate the investigation of \llms' robustness on prompts, 
\rthree{\method integrates $4$ types of attacks: (1) character-level attacks~\citep{textbugger, deepwordbug}, which manipulate texts by
introducing typos or errors to words; (2) word-level attacks~\citep{textfooler, bertattack}, which aim to replace words with synonyms or
contextually similar words to deceive LLMs; (3) sentence-level attacks~\citep{checklist, stresstest}, which append irrelevant or extraneous
sentences to the end of prompts, intending to distract LLMs; (4) semantic-level attacks~\citep{zhu2023promptbench}, which  simulate the linguistic behavior of people from different countries. Details can be found in Appendix~\ref{append:prompt attacks}}
These attacks can be easily called via the \prompt{prompt\_attack} interface.
It also supports the usage of curated adversarial prompts to efficiently evaluate robustness.

\textbf{Different evaluation protocols.}
By default, \method supports the standard protocol, i.e., the direct inference.
\method further supports dynamic \citep{zhu2023dyval} and semantic \citep{liu2023meta} evaluation protocols by dynamically generating testing data.
It is open to integrate more new protocols to avoid data contamination.

\textbf{Analysis tools.}
Finally, \method offers a series of analysis tools to help researchers analyze their results.
Particularly, it support sweep running to get the benchmark results.
Then, attention visualization analysis can be done through the \prompt{utils} interface.
\method also supports word frequency analysis to analyze the words used in attacks as well as defense analysis by integrating word correction tools.

\subsection{Evaluation pipeline}

\method allows easy construction of an evaluation pipeline via four steps.
Firstly, specify task and then load dataset via \prompt{pb.DatasetLoader}. \method offers a streamlined one-line API for loading the desired dataset.
Secondly, users can customize \llms using the \prompt{pb.LLMModel}, which provides integrated inference pipelines compatible with most \llms implemented in Huggingface.
Thirdly, the prompt for the specified dataset and task is defined via \prompt{pb.Prompt}.
Users have the option to input a list of prompts for evaluation and performance comparison. In cases where no prompts are supplied, our default prompts for the dataset are utilized.
Finally, the pipeline requires the definition of input and output processing functions via \prompt{class InputProcess} and \prompt{class OutputProcess} defined in \prompt{pb.utils.dataprocess}, as well as the evaluation function via \prompt{pb.metrics}.
The detailed introduction of the components are shown in Appendix~\ref{sec-append-detail}.

\subsection{Supported research topics}

\rthree{Compared to current evaluation libraries, \method is designed mainly for research purpose, thus it is easy to customize for different topics.}
As shown in \figurename~\ref{fig-main}(b), it supports different evaluation topics from the research community including benchmarks, scenarios, and protocols.
In benchmarks research, it supports standard natural language understanding, natural language generation, and reasoning tasks. It can also be extended to support research on AI agent and interdisciplinary study.
In scenario research, it supports adversarial and out-of-distribution evaluation, and can also support other topics such as hallucination and bias by changing the \prompt{metrics} and \prompt{DatasetLoader} interface.
In protocol, it naturally supports standard and dynamic evaluation, and can further be verified by including measurement theory.
\method offers three leaderboards to allow easy comparison: adversarial prompt attack, prompt engineering, and dynamic evaluation, as shown in Appendix~\ref{sec-append-benchmark}.
Researchers are welcome to submit new results to our platform.
Extensibility is shown in Appendix~\ref{sec-append-exten} that allows convenient extension of the framework.

\section{Conclusion and Discussion}

We presented \method, a unified framework for \llms evaluation, designed in a modular fashion to build evaluation pipelines with various models, tasks, and prompts. It supports research in prompt engineering, adversarial attacks, and dynamic evaluation. \method is the first step in assessing and exploring the capabilities of current \llms. We believe our benchmark and analysis will inform the design of more robust, human-aligned models. \rthree{In the future, new datasets, evaluation protocols, prompt variations, and analytical tools will be added into \method. We also welcome any contributions for \method.

Despite its versatility, \method has limitations. It may not cover all evaluation scenarios, and some metrics might miss nuanced performance differences. The framework's effectiveness depends on the quality and diversity of datasets and prompts. Addressing these limitations is a priority for our ongoing and future work.
}

\bibliography{ref}

\newpage

\appendix

\section{Comparison with Related Code Libraries}
\label{sec-append-related}

\begin{table}[htbp]
\caption{Comparison with related code libraries.}
\label{tb-related-libraries}
\centering
\resizebox{\textwidth}{!}{
\begin{tabular}{ p{4.5cm} p{5cm} p{3.5cm} p{4cm} }
\toprule
\textbf{Library} & \textbf{Purpose} & \textbf{Customization} & \textbf{Functions} \\
\midrule
\textbf{\makecell[tl]{OpenAI Evals \\ \citep{2023evals}}} & A framework for evaluating LLMs or systems built using LLMs. & model, dataset, prompt, eval & \makecell[tl]{Evaluation pipelines\\Benchmarks} \\
\hline
\textbf{\makecell[tl]{OpenCompass\\ \citep{2023opencompass}}} & One-stop platform for large model evaluation, aiming to provide a transparent benchmark for foundation models. & model, dataset & \makecell[tl]{Evaluation pipelines\\Benchmarks\\Leaderboards} \\
\hline
\textbf{\makecell[tl]{promptfoo\\ \citep{2023promptfoo}}} & A framework for evaluating prompts and large language models. & model, dataset, prompt, eval & Evaluation pipelines \\
\hline
\textbf{\makecell[tl]{LM Evaluation Harness\\ \citep{eval-harness}}} & A framework for evaluation of autoregressive LMs. & model, dataset, prompt, eval & \makecell[tl]{Evaluation pipelines\\Benchmarks\\Leaderboards} \\
\hline
\textbf{\makecell[tl]{HELM\\ \citep{liang2022holistic}}} & Holistic Evaluation of Language Models. & model & \makecell[tl]{Evaluation pipelines\\Benchmark\\Leaderboard} \\
\hline
\textbf{PromptBench (Ours)} & Research-focused evaluation toolkit. & model, dataset, prompt, prompt engineering, eval & \makecell[tl]{Evaluation pipelines \\ Prompt Engineering \\ Prompt attacks \\ Dynamic evaluation \\ Leaderboard} \\
\hline
\end{tabular}
}
\end{table}

\section{Details of \method}
\label{sec-append-detail}



\subsection{Models}
\label{sec:appendix-model}

In this section, we list the LLMs and VLMS implemented in \method.


\paragraph{Open-source LLMs:}
\begin{itemize}[leftmargin=1em]
\setlength\itemsep{0em}
    \item \textbf{Flan-T5-large \citep{t5}}: Google's Flan-T5-large, a variation of the Text-to-Text Transfer Transformer (T5).
    \item \textbf{Dolly-6B \citep{dolly}}: The Dolly-6B model, developed by Databricks, is a 6-billion parameter causal language model. It is an extension of EleutherAI’s GPT-J \citep{gpt-j}, further refined with Stanford's Alpaca \citep{alpaca} dataset comprising 52K question/answer pairs.
    \item \textbf{Vicuna series \citep{vicuna}}: Developed from the LLaMA-13B base model, Vicuna-13B integrates over 70K user-shared conversations from ShareGPT.com, leveraging public APIs for data acquisition.
    \item \textbf{Cerebras series \citep{cerebras}}: Modeled on the GPT-3 architecture, Cerebras-13B is part of the Cerebras-GPT series, trained according to Chinchilla scaling laws \citep{hoffmann2022training} to optimize computational efficiency.
    \item \textbf{Llama2 series \citep{llama2}}: Engineered by Meta AI's FAIR team, the Llama2 model is an autoregressive language model adopting the transformer architecture.
    \item \textbf{GPT-NEOX-20B \citep{neox}}: This variant, part of the extensive GPT model series, features 20 billion parameters, exemplifying large-scale language model implementation.
    \item \textbf{Flan-UL2 \citep{ul2}}: Flan-UL2, an encoder-decoder model, is grounded in the T5 architecture and enhanced with Flan prompt tuning and dataset techniques.
    \item \textbf{phi-1.5 and phi-2 \citep{textbooks2}}: phi-1.5 is an LLM with 1.3 billion parameters, builds upon the dataset used for phi-1 with the addition of diverse NLP synthetic texts.
    \item \textbf{Mistral 7B \citep{jiang2023mistral}}: Mistral 7B is trained by Mistral AI team. It excels in tasks like reasoning, mathematics, and code generation. It uses grouped-query attention for faster inference and sliding window attention for efficient handling of sequences. There's also an instruction-following version, Mistral 7B-Instruct.
    \item \textbf{Mixtral8x7B \citep{mixtral}}: Engineering by Mistral AI team, this model is a high-quality sparse mixture of experts model (SMoE) with open weights. There's also an instruction-following version, Mixtral 8x7B Instruct.
    \item \textbf{Baichuan2 series \citep{yang2023baichuan}}: Baichuan2 is developed by Baichuan Intelligent. Trained on 2.6 trillion high-quality tokens, it achieves the best results in its size class on multiple authoritative benchmarks in Chinese, English, and multilingual general and domain-specific tasks.
    \item \textbf{Yi series \citep{yi}}: Developed by 01.AI, the Yi series are next-generation open-source large language models. Trained on a 3T multilingual corpus, they excel in language understanding, commonsense reasoning, and reading comprehension.
    
    \ronertwo{
    \item \textbf{BLIP2 \citep{li2023blip}}: This visual-language model is proposed by Junnan Li, Dongxu Li, Silvio Savarese, Steven Hoi. BLIP-2 leverages frozen pre-trained image encoders and large language models (LLMs) by training a lightweight, 12-layer Transformer encoder in between them, achieving excellent performance in various vision-language tasks
    \item \textbf{LLaVA \citep{liu2024visual}}: LlaVA (Language-Image LLaMA) is a multimodal model combining language and image data. It extends the LLaMA architecture to handle both modalities, enabling tasks like image captioning, visual question answering, and image-based text generation.
    \item \textbf{Qwen-VL series\citep{bai2023qwen}}: Qwen-VL (Qwen Large Vision Language Model) is the multimodal version of the large model series, Qwen (abbr. Tongyi Qianwen), proposed by Alibaba Cloud. Qwen-VL accepts image, text, and bounding box as inputs, outputs text, and bounding box.
    \item \textbf{InternLM-XComposer2-VL \citep{internlmxcomposer2}}: InternLM-XComposer2 is a cutting-edge vision-language model excelling in free-form text-image composition and comprehension, crafting content from diverse inputs like outlines, detailed specs, and reference images. Using a Partial LoRA (PLoRA) approach, it balances vision understanding and text composition.
    }
\end{itemize}

\paragraph{Proprietary LLMs:}
\begin{itemize}[leftmargin=1em]
\setlength\itemsep{0em}
    \item \textbf{ChatGPT \citep{chatgpt} and GPT-4 \citep{openai2023gpt4}}: OpenAI's ChatGPT and GPT-4 are advanced iterations of the GPT series. ChatGPT is tailored for interactive tasks, while GPT-4 is the most proficient in the series and supports image input.
    \item \textbf{PaLM 2 \citep{anil2023palm}}: PaLM 2 is an advanced language model that excels in multilingual and reasoning capabilities, offering greater computational efficiency than its predecessor, PaLM. This Transformer-based model enhances performance across various model sizes in English, multilingual tasks, and reasoning challenges.
    
    \ronertwo{
    \item \textbf{Gemini \citep{gemini}}: The Gemini model is a multimodal language model developed by Google AI, capable of extracting insights from a diverse array of data formats, including images, and video.
    }
\end{itemize}

\subsection{Tasks and Datasets}
\label{sec:appendix-task}

\begin{itemize}[leftmargin=1em]
\setlength\itemsep{0em}

\item \textbf{GLUE \citep{wang2019glue}}: The GLUE benchmark (General Language Understanding Evaluation)  offers a suite of tasks to evaluate the capability of NLP models in understanding language. For this research, we employed 8 specific tasks: Sentiment Analysis (\sst \citep{sst2}), Grammar Correctness (\cola \citep{cola}), Identifying Duplicate Sentences (\qqp \citep{qqp}, \mrpc \citep{mrpc}), and various Natural Language Inference tasks (\mnli \citep{mnli}, \qnli \citep{wang2019glue}, \rte \citep{wang2019glue}, \wnli \citep{wnli}).

\item \textbf{MMLU \citep{mmlu}}: The MMLU dataset tests the broad knowledge and problem-solving skills of large language models through 57 tasks with multiple-choice questions in fields like mathematics, history, and computer science. It is a comprehensive multitask benchmark.

\item \textbf{SQuAD V2 \citep{squad}}: The SQuAD v2 dataset is pivotal in training and assessing NLP models for reading comprehension. It builds upon the original SQuAD by adding unanswerable questions, making it more challenging. Models must either identify the correct answer in the text or recognize questions as unanswerable.

\item \textbf{UN Multi \citep{multiun}}: Comprising texts in the six official United Nations languages, the Multi UN dataset is a vast parallel corpus from UN documents. However, its focus on formal texts may restrict its use in informal or conversational language contexts.

\item \textbf{IWSLT 2017 \citep{iwslt}}: Designed for spoken language translation system evaluation, the IWSLT 2017 dataset includes multilingual, multi-domain text data, primarily from the TED Talks Open Translation Project. It encompasses numerous language pairs, providing a rich resource for translation tasks.

\item \textbf{Math \citep{math}}: The DeepMind Mathematics Dataset assesses AI models' mathematical reasoning by posing a wide array of math problems, from algebra to calculus. It tests the models' understanding and logical reasoning in mathematics.

\item \textbf{BIG-Bench \citep{bigbench}}: BIG-bench is a collaborative benchmark designed to evaluate the capabilities of large language models and predict their future potential. It consists of over 200 tasks, contributed by 444 authors from 132 institutions, covering a wide range of topics like linguistics, math, common-sense reasoning, and more. These tasks are intended to probe areas believed to be beyond the current capabilities of LMs. 

\item \textbf{GSM8K \citep{cobbe2021gsm8k}}: The GSM8K dataset is a collection of 8.5K high-quality, linguistically diverse grade school math word problems. It was created by human problem writers and is divided into 7.5K training problems and 1K test problems. These problems, which require 2 to 8 steps to solve, primarily involve basic arithmetic operations and are designed to be solvable by a bright middle school student. 

\item \textbf{CommonsenseQA \citep{talmor2019commonsenseqa}}: The CommonsenseQA dataset is a challenging commonsense question-answering dataset. It comprises 12,247 questions with 5 multiple-choice answers each. 

\item \textbf{QASC \citep{khot2020qasc}}: QASC (Question Answering via Sentence Composition) is a specialized collection designed for question-answering tasks with a focus on sentence composition. It comprises 9,980 eight-way multiple-choice questions about grade school science, divided into 8,134 for training, 926 for development, and 920 for testing .(In our evaluation, we use development part.) The dataset is notable for its emphasis on multi-hop reasoning, requiring the retrieval and composition of facts from a broad corpus to answer each question.

\item \textbf{NummerSense \citep{lin2020birds}}: NumerSense is a unique numerical commonsense reasoning probing task, featuring a diagnostic dataset with 3,145 masked-word-prediction probes. This dataset has applications in tasks such as knowledge base completion and open-domain question answering.

\ronertwo{

\item \textbf{VQAv2 \citep{goyal2017making}}: Visual Question Answering (VQA) v2.0 is a dataset containing open-ended questions about images. These questions require an understanding of vision, language and commonsense knowledge to answer. It is the second version of the VQA dataset.

\item \textbf{NoCaps \citep{agrawal2019nocaps}}: NoCaps is a benchmark dataset for image captioning models that can describe images containing novel objects from object detection datasets. It consists of 166,100 human-generated captions describing 15,100 images from the Open Images validation and test sets.

\item \textbf{MMMU \citep{yue2023mmmu}}: MMMU is a comprehensive benchmark designed to evaluate multimodal models on massive multi-discipline tasks demanding college-level subject knowledge and deliberate reasoning. MMMU includes 11.5K meticulously collected multimodal questions from college exams, quizzes, and textbooks. These questions span 30 subjects and 183 subfields, comprising 32 highly heterogeneous image types, such as charts, diagrams, maps, tables, music sheets, and chemical structures.

\item \textbf{MathVista \citep{lu2023mathvista}}: MathVista is a comprehensive benchmark for mathematical reasoning in visual contexts. It includes three new datasets: IQTest, FunctionQA, and PaperQA. These datasets cover various visual domains and are designed to evaluate logical reasoning on puzzle test figures, algebraic reasoning using functional plots, and scientific reasoning with academic paper figures, respectively.

\item \textbf{AI2D \citep{kembhavi2016diagram}}: AI2D is a dataset of illustrative diagrams for research on diagram understanding and associated question answering. It contains 5000 grade-school science diagrams with over 150,000 rich annotations, their ground truth syntactic parses, and more than 15,000 corresponding multiple-choice questions.

\item \textbf{ChartQA \citep{masry-etal-2022-chartqa}}: ChartQA is a large-scale dataset of complex reasoning questions over charts that involve visual and logical operations, covering 9.6K human-written questions as well as 23.1K questions generated from human-written chart summaries.

\item \textbf{ScienceQA \citep{lu2022learn}}: ScienceQA is a large-scale dataset for multimodal reasoning with diverse science topics and annotations of answers, lectures and explanations. It covers 26 topics, 127 categories and 379 skills from natural science, language science, and social science and so on.
}

\end{itemize}

\subsection{Evaluation protocols}
DyVal \citep{zhu2023dyval} is an approach for dynamic evaluation of \llms by creating complexity-tailored evaluation samples on-the-fly, as opposed to relying on static benchmarks. DyVal synthesized seven distinct reasoning tasks, including: (1) Mathematics, focusing on arithmetic calculations and linear equation solving; (2) Logical Reasoning, involving boolean, deductive, and abductive logic; and (3) Algorithmic Analysis, covering reachability and the maximum sum path problem.
MSTemp~\citep{liu2023meta} stands for the semantic evalaution protocol which generate out-of-distribution samples by relying on evlauator LLMs and word replacement.

\subsection{Prompts}
\label{sec:appendix-prompt}

\subsubsection{Prompts}
Our study examines four prompt categories, differentiated by their intended function and the required number of labeled samples. Task-oriented prompts are designed to clearly define the model's task, prompting it to generate outputs relevant to the task using its inherent pre-training knowledge. In contrast, role-oriented prompts position the model as a particular entity, like an expert, advisor, or translator, thereby subtly guiding the expected output format and behavior through the assumed role. Both categories can be adapted for zero-shot and few-shot learning contexts. We randomly choose three training set examples for each task to form the few shot examples. Examples of various prompt types are illustrated in \tablename~\ref{tb-prompt-examples}.

\begin{table}[htbp]
\caption{Examples of $4$ types of prompts.}
\label{tb-prompt-examples}
\centering
\resizebox{\textwidth}{!}{
\begin{tabular}{ c  c m{16cm}}
\toprule

\multirow{4}{*}{\makecell{Zero\\shot}}  & \makecell{Task\\oriented} & { \prompt{Determine if the given pair of statements can be considered the same by responding with 'equivalent' or 'not\_equivalent'.} } \\

\cmidrule(lr){2-3}

& \makecell{Role\\oriented} & { \prompt{As an instrument for question comparison evaluation, consider the questions and determine if their meaning is the same, responding with 'equivalent' for similar questions or 'not\_equivalent' for different questions.} } \\

\midrule

\multirow{7}{*}{\makecell{Few\\shot}}  & \makecell{Task\\oriented} & { \prompt{Review the sentence below and identify whether its grammar is 'Acceptable' or 'Unacceptable': Here are three examples. Sentence: Our friends won't buy this analysis, let alone the next one we propose. Answer: acceptable. Sentence: One more pseudo generalization and I'm giving up. Answer: acceptable. Sentence: They drank the pub. Answer: unacceptable. } }\\

\cmidrule(lr){2-3}

& \makecell{Role\\oriented} &  { \prompt{Functioning as a grammar evaluation tool, analyze the given sentence and decide if it is grammatically correct, responding with 'acceptable' or 'unacceptable': Here are three examples. Sentence: Our friends won't buy this analysis, let alone the next one we propose. Answer: acceptable. Sentence: One more pseudo generalization and I'm giving up. Answer: acceptable. Sentence: They drank the pub. Answer: unacceptable. } }\\

\bottomrule
\end{tabular}
}
\end{table}

\subsubsection{Prompt Engineering}
\label{append:pe}
Prompt engineering is a process of structuring and optimizing prompts to efficiently use AI models. Methods in prompt engineering , such as chain-of-thought \citep{wei2023chainofthought}, generated knowledge prompting \citep{liu2022generated} and so on, help improve reasoning ability and task performance of AI models. We implement 6 prominent prompt engineering methods:
\begin{itemize}[leftmargin=1em]
\setlength\itemsep{0em}
    \item \textbf{Chain-of-Thought \citep{wei2023chainofthought}}: This method involves breaking down complex, multi-step problems into smaller, intermediate steps, enabling Models to tackle more intricate reasoning tasks. Chain-of-Thought differs from standard few-shot prompting by not just providing questions and answers but prompting the model to produce intermediate reasoning steps before arriving at the final answer. 
    \item \textbf{Zero-Shot Chain-of-Thought \citep{kojima2022large}}: Zero-Shot Chain of Thought improves Chain of Thought by simplifying the prompting process. The key innovation in Zero-Shot Chain-of-Thought is appending the phrase ``\prompt{Let's think step by step}'' to the end of a question.
    \item \textbf{EmotionPrompt \citep{li2023large}}: Drawing inspiration from psychology and social science theories about human emotional intelligence, this method adds emotional stimuli to origin prompts. For example: ``\prompt{This is very important to my career}.''
    \item \textbf{Expert Prompting \citep{xu2023expertprompting}}: The key idea is to let model be an expert in role playing. To generate the expert identity, we first provide several instruction-expert pair exemplars, then the model generates an expert identity of this question. Finally, we ask the model to answer the instruction conditioned on expert identity.
    \item \textbf{Generated Knowledge \citep{liu2022generated}}: Generated Knowledge Prompting is a method where a model first generates knowledge and then uses this generated information as additional input to answer questions. It enhances commonsense reasoning in AI without requiring task-specific knowledge integration or a structured knowledge base. 
    \item \textbf{Least to Most \citep{zhou2023leasttomost}}: Least to Most breaks down a complex problem into a series of simpler subproblems and then solves them in sequence. The key idea is to solve each subproblem by using the answers to previously solved subproblems. This method is particularly useful for tasks that require solving problems harder than the exemplars shown in the prompts.
\end{itemize}

Note that there are plenty of prompt engineering techniques and we tried our best to include those general techniques instead of specific prompt engineering techniques such as Tree of Thoughts~\citep{yao2023tree} that requires specific prompt design and decomposition of each problem.

\subsubsection{Adversarial Prompt Attacks}
\label{append:prompt attacks}
Adversarial prompt attacks, as proposed by \cite{zhu2023promptbench}, aims to \emph{simulate} potential disturbances that could naturally arise in practical scenarios. The proposed prompt attacks are intended to resemble common user errors or expressions, as users often make various mistakes when inputting prompts, such as typos, diverse word choices, and different sentence constructions. The prompt attacks encompass four distinct levels:
\begin{itemize}[leftmargin=1em]
\setlength\itemsep{0em}
    \item \textbf{Character-level:} Techniques such as TextBugger~\citep{textbugger} and DeepWordBug~\citep{deepwordbug} are employed. These methods introduce errors or typos into words by altering characters.
    \item \textbf{Word-level:} Attacks like BertAttack~\citep{bertattack} and TextFooler~\citep{textfooler} are utilized. They focus on substituting words with their synonyms or contextually similar alternatives.
    \item \textbf{Sentence-level:}  StressTest~\citep{stresstest} and CheckList~\citep{checklist} are applied. These attacks add irrelevant or redundant sentences to prompts.
    \item \textbf{Semantic-level:} To simulate the linguistic styles of different global regions.
\end{itemize}

\subsection{Pipeline}

\begin{figure}[htbp]
    \centering
    \includegraphics[width=1.0\textwidth]{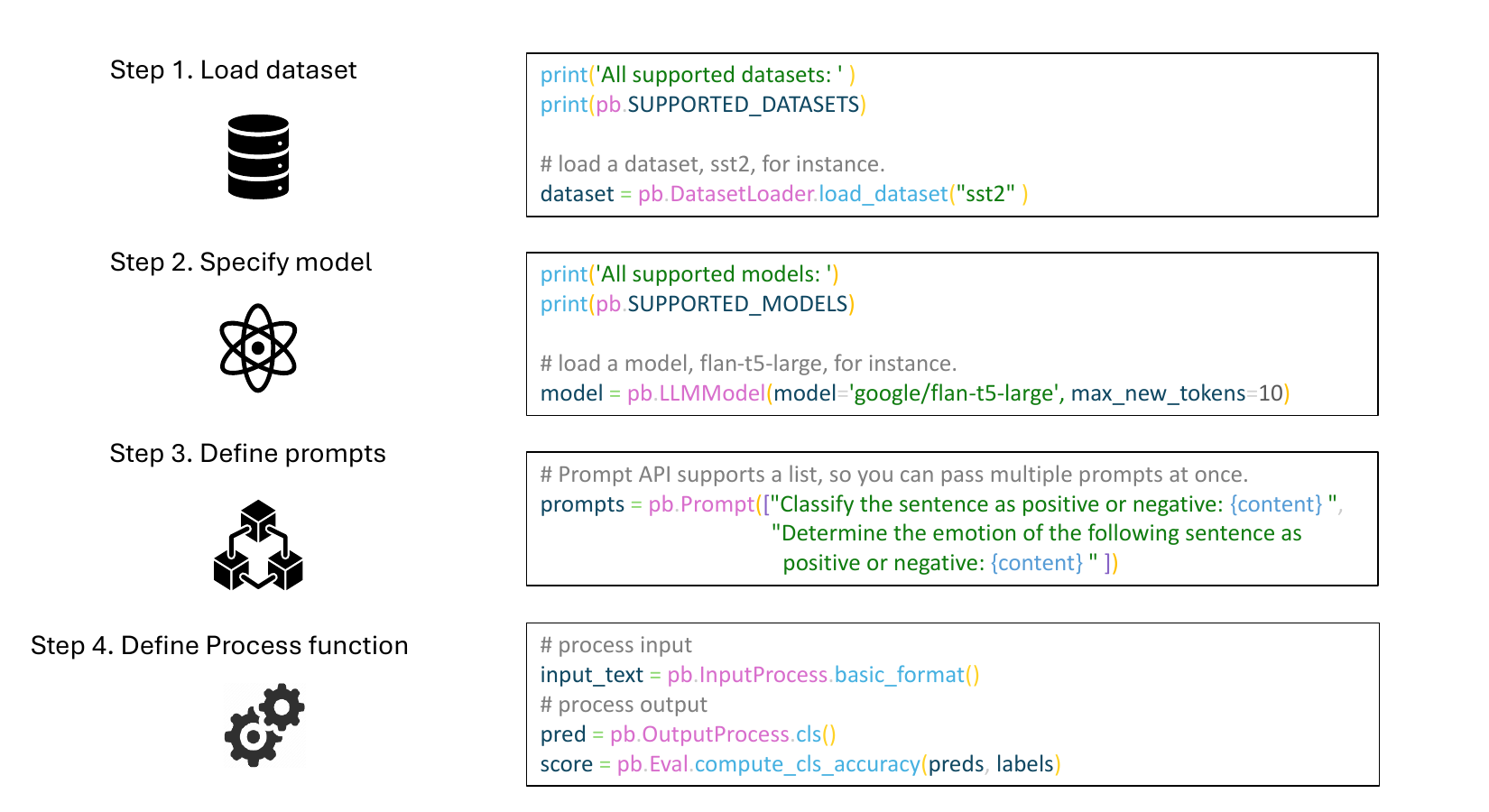}
    \caption{A pipeline for evaluation of \llms.}
    \label{fig-pipeline}
\end{figure}

The full pipeline of using \method for evaluation is shown in \figurename~\ref{fig-pipeline}.

\section{Benchmark Results}
\label{sec-append-benchmark}

\subsection{Adversarial prompt robustness}
\begin{figure}[t!]
    \centering
    \includegraphics[width=0.8\textwidth]{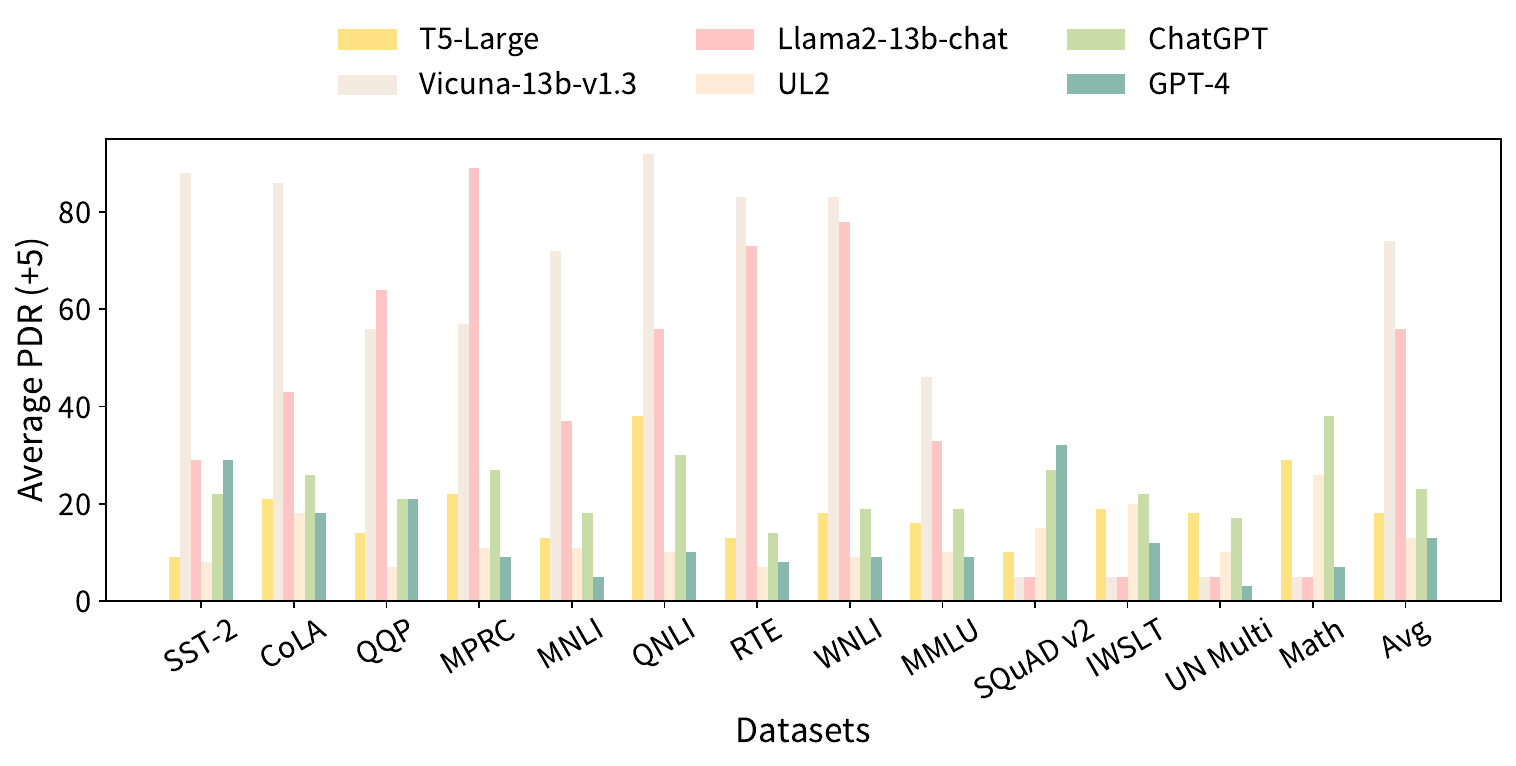}
    \vspace{-.2in}
    \caption{Adversarial prompt robustness results.}
    \label{fig-adv-prompts}
\end{figure}

The partial results of the robustness of different models on a range of tasks are presented in \figurename~\ref{fig-adv-prompts}. All models exhibit vulnerability to adversarial prompts, with ChatGPT and GPT-4 demonstrating the strongest robustness.

\begin{figure}[t!]
    \centering
    \includegraphics[width=0.8\textwidth]{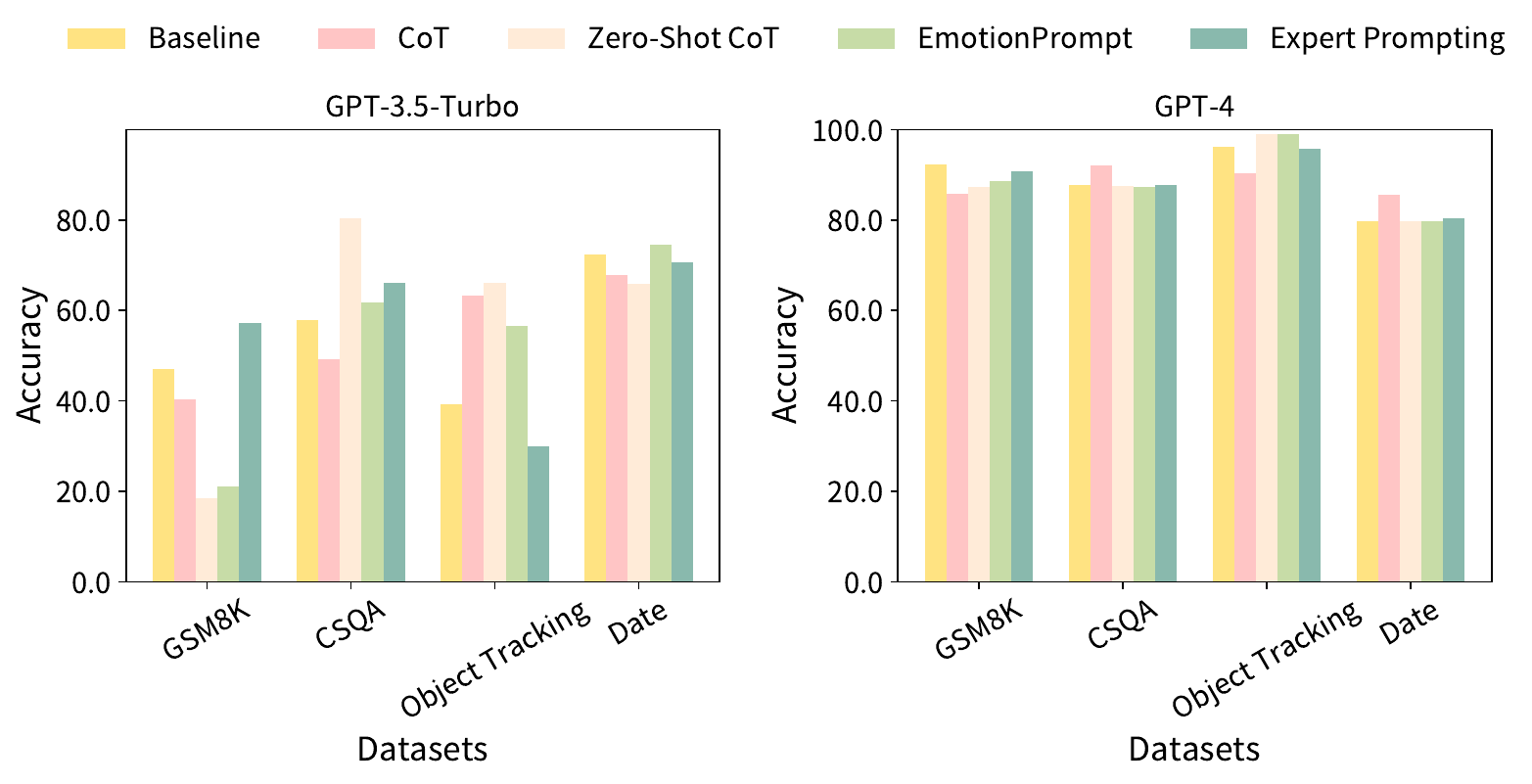}
    \vspace{-.2in}
    \caption{Comparison among different prompt engineering techniques.}
    \label{fig-result-pe}
    
\end{figure}

\begin{figure}[t!]
    \centering
    \includegraphics[width=0.8\textwidth]{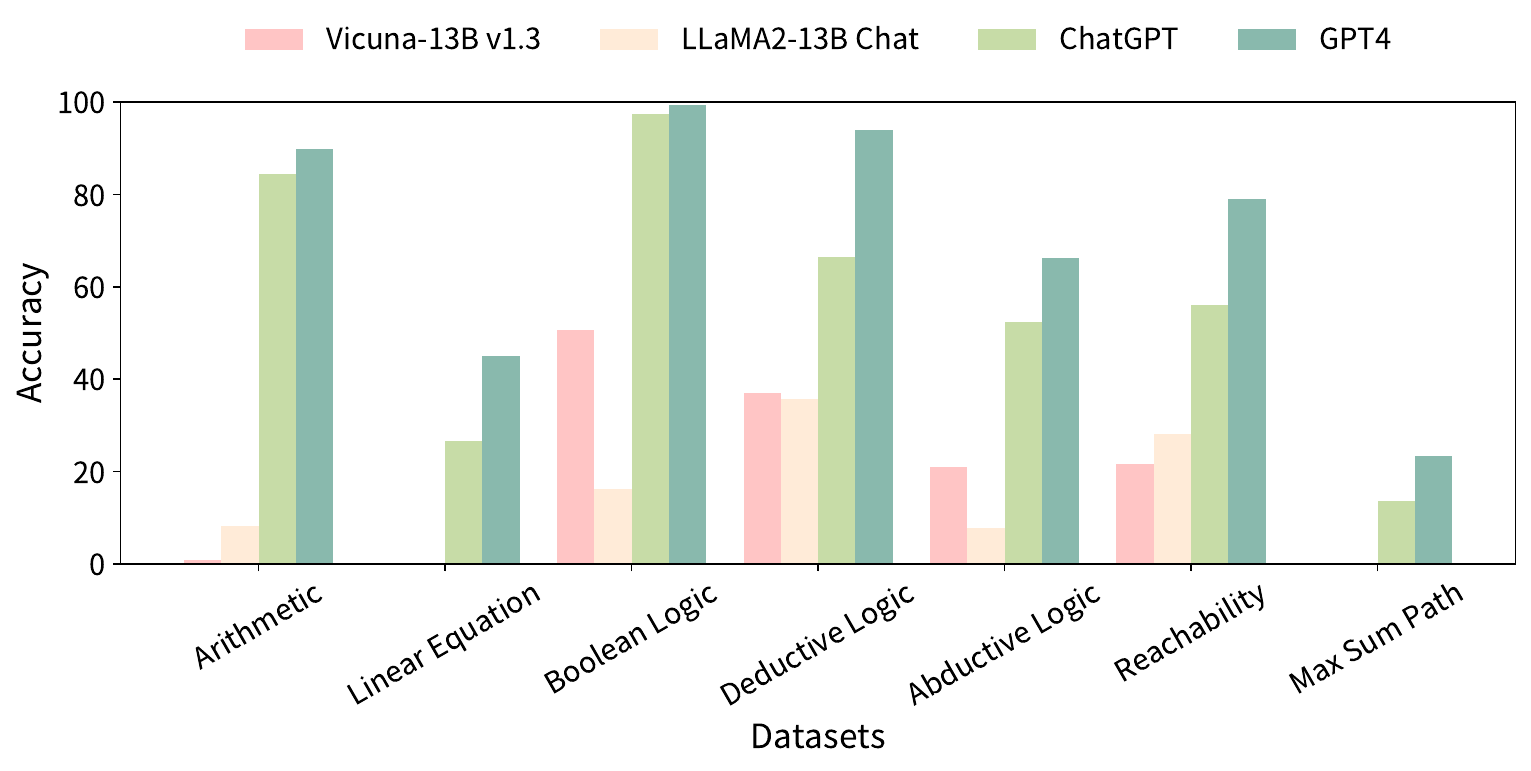}
    \vspace{-.2in}
    \caption{DyVal results.}
    \label{fig-dyval}
\end{figure}

\subsection{Prompt engineering}

Prompt engineering results are shown in \figurename~\ref{fig-result-pe}. Most methods are effective for special fields, so these methods can not surpass the baseline in every dataset.

\subsection{Dynamic evaluation}
\figurename~\ref{fig-dyval} illustrates the outcomes of dynamic evaluations across various models and tasks. GPT-4 outperforms its counterparts significantly, yet there remains potential for enhancement in the performance of linear equation, abductive logic, and max sum path task.

\section{Extensibility}
\label{sec-append-exten}
Each module in \method can be easily extended. In the following, we provide basic guidelines for customizing your own datasets, models, prompt engineering methods, and evaluation metrics. \rtwo{The sample code for adding new datasets, models can be found in \url{https://github.com/microsoft/promptbench/blob/main/examples/add_new_modules.md}}

\subsection{Add new datasets}
Adding new datasets involves two steps:
\begin{enumerate}
    \item Implementing a New Dataset Class: Datasets are supposed to be implemented in \prompt{dataload/dataset.py} and inherit from the \prompt{Dataset} class. For your custom dataset, implement the \prompt{\_\_init\_\_} method to load your dataset. We recommend organizing your data samples as dictionaries to facilitate the input process.
    \item Adding an Interface: After customizing the dataset class, register it in the \prompt{DataLoader} class within \prompt{dataload.py}.
\end{enumerate}

\subsection{Add new models}
Similar to adding new datasets, the addition of new models also consists of two steps.
\begin{enumerate}
    \item Implementing a New Model Class: Models should be implemented in \prompt{models/model.py}, inheriting from the \prompt{LLMModel} class. In your customized model, you should implement \prompt{self.tokenizer} and \prompt{self.model}. You may also customize your own \prompt{predict} function for inference. If the \prompt{predict} function is not customized, the default \prompt{predict} function inherited from \prompt{LLMModel} will be used.
    \item Adding an Interface: After customizing the model class, register it in the \prompt{\_create\_model} function within the \prompt{class LLMModel} and \prompt{MODEL\_LIST} dictionary in \prompt{\_\_init\_\_.py}.
\end{enumerate}

\subsection{Add new prompt engineering methods}
Adding new methods in prompt engineering is similar to steps of C.1 and C.2.
\begin{enumerate}
    \item Implementing a New Methods Class: Methods should be implemented in \\ \prompt{prompt\_engineering} Module. Firstly, create a new \prompt{.py} file for your methods.
    Then implement two functions: \prompt{\_\_init\_\_} and \prompt{query}. For unified management, two points need be noticed: 1. all methods should inherits from \prompt{Base} class that has common code for prompt engineering methods. 2. prompts used in methods should be stored in \prompt{prompts/method\_oriented.py}.
    \item Adding an Interface: After implementing a new methods, register it in the \prompt{METHOD\_MAP} that is used to map method names to their corresponding class.
\end{enumerate}

\subsection{Add new metrics and input/output process functions}
New evaluation metrics should be implemented as static functions in \prompt{class Eval} within the \prompt{metrics} module. Similarly, new input/output process functions should be implemented as static functions in \prompt{class InputProcess} and \prompt{class OutputProcess} in the \prompt{utils} module.


\label{sec:appendix-analysis}

\end{document}